## ROBUST INFERENCE POLICIES: PRELIMINARY REPORT[1]

**Paul E. Lehner**
George Mason University
4400 University Drive
Fairfax, VA 22030
(703) 323-4355
plehner@gmuvax2.gmu.edu

*Abstract* - A series of monte carlo studies were performed to assess the extent to which different inference procedures robustly output reasonable belief values in the context of increasing levels of judgmental imprecision. It was found that, when compared to an equal-weights linear model, the Bayesian procedures are more likely to deduce strong support for a hypothesis. But, the Bayesian procedures are also more likely to strongly support the <u>wrong</u> hypothesis. Bayesian techniques are more powerful, but are also more error prone.

### 1.0 INTRODUCTION

Reasoning under uncertainty often involves a great deal of judgmental imprecision. The subjective (or data base retrieved) uncertainty estimates that serve as the ingredients of an uncertainty calculus are often perceived as arbitrary, imprecise or uncertain. One consequence of this judgmental imprecision is that many decision makers (and researchers) avoid using an explicit uncertainty calculus for fear of being subject to a garbage in-garbage out problem. For instance, in applied decision analysis one often finds resistance to using decision analytic models because subjective assessments are viewed with suspicion.

This paper examines the garbage in-garbage out problem. In particular, this paper examines the extent to which several different inference procedures are robustly accurate in the context of increasing levels of judgmental imprecision.

### 2.0 METHOD AND RESULTS

Currently, we are engaged in a series of monte carlo simulations. Although the analysis has just begun, the results to date show an interesting and stable pattern when comparing Bayesian techniques with some other simpler inference procedures. The best way to describe the methodology is to present the details of a single monte carlo run. Following this, variations on this prototypical example are discussed.

--------------------

1. The research was supported in part by a grant from the National Science Foundation under Grant NSF-EET-8820124 and by the Center for Excellence in Command, Control, Communications, and Intelligence at George Mason University. The Center's general research program is sponsored by the Virginia Center for Innovative Technology, MITRE Corporation, the Defense Communications Agency, CECOM, PRC/ATI, ASD(C3I), TRW, AFCEA, and AFCEA NOVA.

## 2.1 The Prototypical Example: Method

This run involved an inference problem with one hypothesis (H) node and four evidence nodes (A-D). Each node is bi-valued. A <u>true probability distribution</u> was defined by randomly assigning (from 0-1 uniform distribution) a value to each instance of $P(H=T)$, $P(A=T|H)$, $P(B=T|HA)$, $P(C=T|HAB)$, and $P(D=T|HABC)$.[2] In addition, an <u>underlying distribution of belief values</u> was also assigned by taking each of the probability statements used to define the probability distribution and adding some random error. For instance, for $P(B=T|H=F,A=T)$ we would set

$$B(B=T|H=F,A=T) = P(B=T|H=F,A=T) + error.$$

For this run the error distribution was a uniform. Specifically,

$$B(x|y) = min + (max-min)*RND,$$

where

$$min = maximum[.00001, P(x|y)-(range/2)]$$
$$max = minimum[.99999, P(x|y)+(range/2)].$$

Here $P(x|y)$ is the true probability, $B(x|y)$ is the belief value or assessed probability, RND is a random number and "range" is the range of the error distribution.

Once calculated, the belief values defined a fully-specified and coherent distribution over the possible states. The inputs to each of the alternative inference procedures were derived from this underlying distribution of belief values.

For each inference procedure and possible evidential state (combination of values for A-D) the <u>relative belief</u> (RB) for H=T vs. H=F was calculated as follows. For each evidential state, relative belief corresponds to the updated or posterior belief value that would be generated by each inference procedure.

<u>Proper Bayes</u>. For each evidential state, relative belief was calculated by

$$RB(H=T|ABCD) = \{B(H=T \& ABCD)/[B(H=T \& ABCD) + B(H=F \& ABCD)]\}.$$

<u>Naive Bayes</u>. For each evidential state, relative belief was calculated by

$$RB(H=T|ABCD) = \frac{B(H=T)*B(A|H=T)*B(B|H=T)*B(C|H=T)*B(D|H=T)}{B(H=T)*B(A|H=T)*B(B|H=T)*B(C|H=T)*B(D|H=T) + B(H=F)*B(A|H=F)*B(B|H=F)*B(C|H=F)*B(D|H=F)}$$

----------------------

2. A notational note. An expression such as $P(A=T|H=T\&BC)$ is a template for a set of expressions. In this case, $P(A=T|H=T,B=T,C=T)$, $P(A=T|H=T,B=T,C=F)$, ... are instances of this template.

<u>Strong Naive Bayes</u>. This was calculated in the same way as Naive Bayes with the exception that an evidence item x was dropped from the calculation whenever $2/3 < B(x|H=T)/B(x|H=F) < 3/2$.

<u>Simple Linear</u>. This technique calculates relative belief by simply adds pros and cons. The calculation proceeds as follows. Start with SL=0.

(S1)
$$\text{If } B(H=T) > .5 \text{ then } SL = SL + 1$$
$$\text{else if } B(H=T) < .5 \text{ then } SL = SL - 1.$$

For each evidence item x, let $LR(x|H) = B(x|H=T)/B(x|H=F)$.

(S2)
$$\text{If } LR(x|H) > 1 \text{ then } SL = SL + 1$$
$$\text{else if } LR(x|H) < 1 \text{ then } SL = SL - 1.$$

Finally, normalize by setting $RB(H=T|ABCD) = (SL+5)/10$.

<u>Strong Linear</u>. This was exactly the same as Simple Linear except that S1 and S2 were replaced by

$$\text{If } B(H=T) > .7 \text{ then } SL = SL + 1$$
$$\text{else if } B(H=T) < .3 \text{ then } SL = SL - 1, \text{ and}$$

$$\text{If } LR(x|H) > (3/2) \text{ then } SL = SL + 1$$
$$\text{else if } LR(x|H) < (2/3) \text{ then } SL = SL - 1.$$

These inference procedures are representative of some of the more popular point-valued system found in the uncertain reasoning literature. Proper Bayes is often considered to be the normative ideal for uncertain inference (e.g., Pearl, 1988). Bayesian approaches with conditional independence assumptions are often used to approximate Proper Bayes. Simple Linear models are often recommended, because they are often more accurate that human experts (e.g., Dawes, 1979).

Three error measures were used. The first is mean-squared error (MSE), sometimes called a Brier score. This measure is of interest because MSE is a nondistorting scoring rule that can be used to derive the probability calculus as a normative calculus of belief. Expected-MSE is easily calculated since we know (a) the true probability of each evidential state, (b) the true probability of H=T given each evidential state, and (c) the relative belief for H=T given that evidential state.

The second measure is Pe/(Pe+Pc), and is called relative error (RE). To calculate RE we first set an upper and lower decision threshold (U and L respectively). For this run, U=.65 and L=.35. Then Pe (probability of error) and Pc (probability of correct decision) are calculated as follows:

$$Pe = P(RB(H=T) > U|H=F)*P(H=F) + P(RB(H=F) < L|H=T)*P(H=T)$$
and
$$Pc = P(RB(H=F) > U|H=F)*P(H=F) + P(RB(H=T) < L|H=T)*P(H=T).$$

Since the true probability of each state is known, expected-RE was also straightforward to calculate.

The third measure is the normalized mean difference between the distribution of relative belief when H=F vs. when H=T. This measure (d′) is commonly used in Signal Detection Theory to measure the sensitivity of a signal detector to the presence of a signal. Here it measures the "sensitivity" of the inference procedure to the presence of the hypothesis. From Lehner and Ulvila (1990) we note that when U and L are symmetric, d′ is approximately $z(1-Pe)+z(Pc)$, where $z(x)$ is the z-score for x.

## 2.2 The Prototypical Example: Results

The following results are based on 1000 cases. Each case involved a different true probability distribution over the possible states.

TABLE 1
EXPECTED MEAN-SQUARED ERROR FOR PROTOTYPICAL EXAMPLE

| Error Range | Simple Linear | Strong Linear | Naive Bayes | Strong Bayes | Proper Bayes | Minimum Possible |
|-----|-----|-----|-----|-----|-----|-----|
| 0.000 | 0.171 | 0.166 | 0.121 | 0.123 | 0.088 | 0.088 |
| 0.200 | 0.175 | 0.169 | 0.127 | 0.129 | 0.094 | 0.086 |
| 0.400 | 0.178 | 0.172 | 0.134 | 0.136 | 0.108 | 0.083 |
| 0.600 | 0.191 | 0.185 | 0.161 | 0.162 | 0.137 | 0.085 |
| 0.800 | 0.202 | 0.199 | 0.186 | 0.188 | 0.163 | 0.083 |
| 1.000 | 0.230 | 0.219 | 0.224 | 0.226 | 0.210 | 0.087 |
| 1.200 | 0.259 | 0.239 | 0.268 | 0.269 | 0.264 | 0.086 |
| 1.400 | 0.276 | 0.253 | 0.300 | 0.300 | 0.303 | 0.085 |
| 1.600 | 0.287 | 0.265 | 0.331 | 0.331 | 0.345 | 0.086 |
| 1.800 | 0.293 | 0.274 | 0.352 | 0.352 | 0.370 | 0.084 |
| 2.000 | 0.296 | 0.276 | 0.361 | 0.361 | 0.387 | 0.088 |

Table 1 shows the expected-MSE for each technique at increasing levels of calibration error (defined by the Error Range). There are several features to note about Table 1. First, there is an extra column. The Minimum Possible column indicates the minimum possible expected-MSE score. Note that when range = 0 (no calibration error) the Proper Bayes technique achieves, as expected, the minimum possible score. Second, as calibration error increases, expected-MSE increases more rapidly for the Bayesian than for the linear techiques. This is a robust result which we've observed on all our runs. Third, it is worth noting that Naive Bayes and Simple Linear, as well as Strong Bayes and Strong Linear use exactly the same input values. Consequently, the differences between these two pairs of techniques are due entirely to how these input values are aggregated into an overall assessment of relative belief.

Finally, .25 and .33 are important thresholds for expected-MSE scores. A MSE of .25 be guaranteed by asserting relative belief =.5 under all conditions. Consequently, according to MSE all the techniques eventually due worse than saying "I don't know." under all circumstances. The Proper Bayes technique, however, is the first to pass this threshold. Similarly, a MSE of .33 can be achieve by randomly assigning relative belief values. All the Bayesian techniques eventually do worse than assigning relative belief values randomly.

At first these results seem exciting. They demonstrate that Bayesian can be outperformed with a simple inference procedure even when using a performance criteria (Expected-MSE) that is used to axiomatically derive Bayesian procedures. A more careful analysis, however, reveals another conclusion. MSE is a poor error measure. The reasons for this will become apparent below.

TABLE 2
EXPECTED-d' RESULTS FOR PROTOTYPICAL EXAMPLE

| Error Range | Simple linear | Strong linear | Naive Bayes | Strong Bayes | Proper Bayes |
|-------|-------|-------|-------|-------|-------|
| 0.000 | 1.37 | 1.56 | 1.90 | 1.88 | 2.32 |
| 0.200 | 1.32 | 1.50 | 1.83 | 1.80 | 2.25 |
| 0.400 | 1.32 | 1.48 | 1.76 | 1.74 | 2.08 |
| 0.600 | 1.13 | 1.25 | 1.48 | 1.46 | 1.76 |
| 0.800 | 1.02 | 1.06 | 1.24 | 1.22 | 1.51 |
| 1.000 | 0.70 | 0.69 | 0.91 | 0.89 | 1.09 |
| 1.200 | 0.41 | 0.48 | 0.56 | 0.56 | 0.72 |
| 1.400 | 0.23 | 0.31 | 0.36 | 0.36 | 0.45 |
| 1.600 | 0.12 | 0.16 | 0.19 | 0.17 | 0.22 |
| 1.800 | 0.06 | 0.03 | 0.11 | 0.10 | 0.11 |
| 2.000 | 0.02 | 0.01 | 0.05 | 0.04 | 0.02 |

Table 2 shows the results for expected-d'. Recall that d' is the normalized difference between means of the distributions of belief values when the hypothesis is true vs. false. This measure tells a very different story than MSE. Here Proper Bayes consistently outperforms the other inference procedures. This contrasts sharply with the MSE results. According to the MSE measure, at Error Range = 1.2 Proper Bayes does worse than asserting "I don't know -- .5" under all conditions. The d' measure, however, still has the two distribution .72 standard deviations apart.

Although sensitivity (d') is a useful measure, it is also an incomplete measure. d' is invariant to linear transformations of the measurement scale. A belief/probability scale, however, has an absolute value interpretation. It is the absolute value on this scale that drives behavior. RE is an error measure that is sensitive to absolute values. Table 3 compares the various inference procedures in terms of expected-RE.

The RE measure tells yet another story. Proper Bayes and Strong Linear consistently outperform the other three methods. Compared with each other, Strong Linear exhibits an arching effect relative to Proper Bayes. At the extremes, Proper Bayes does better than Strong Linear, but for the middle range Strong Linear does consistently better than Proper Bayes. These results suggests that, for the most part, Strong Linear is less likely than Proper Bayes to inappropriately output a belief value beyond a decision threshold.

TABLE 3
EXPECTED-RE RESULTS FOR PROTOTYPICAL EXAMPLE

| Error range | Simple linear | Strong linear | Naive Bayes | Strong Bayes | Proper Bayes |
|-------|-------|-------|-------|-------|-------|
| 0.000 | .126 | .092 | .122 | .125 | .086 |
| 0.200 | .133 | .097 | .131 | .134 | .093 |
| 0.400 | .135 | .101 | .143 | .143 | .110 |
| 0.600 | .169 | .134 | .182 | .182 | .149 |
| 0.800 | .187 | .168 | .218 | .221 | .185 |
| 1.000 | .270 | .224 | .284 | .287 | .258 |
| 1.200 | .359 | .328 | .361 | .363 | .337 |
| 1.400 | .416 | .387 | .411 | .410 | .394 |
| 1.600 | .457 | .444 | .455 | .457 | .448 |
| 1.800 | .480 | .487 | .474 | .476 | .475 |
| 2.000 | .491 | .496 | .489 | .490 | .494 |

By this time the reader may be confused. Using three different error measures three entirely different patterns of result obtain. As it turns out, however, these results have a straightforward explanation. Figure 1 depicts this explanation for the Strong Linear and Proper Bayes techniques. While Proper Bayes is the more sensitive, Strong Linear exhibits less variance in the calculated belief values. As a result, the decision thresholds are farther out on the tails of the Strong Linear distributions than on the Proper Bayes distributions. It should be noted that Figure 1 is also consistent with the result that for Strong Linear $P_e + P_c$ was always much lower (around .4) than for Proper Bayes (around .85).

In general, these results suggest that the Strong Linear method is a relatively safe, albeit weak, inference procedure. This method often returns an intermediate value, but when it does return an extreme value (i.e., strong support) it is less likely to be in error than many other techniques -- even Proper Bayes.

## 2.3 Variations on Prototype Example

The previous section outlined the results for a single run. In this section some alternative monte carlo runs are described. Because of space limitations, it is not possible to show all the results. Instead, I will limit discussion to describing the main differences from the prototypical example.

*Marginalized vs. Direct Assessments* - The prototypical example assumed that inferencing agents were coherent. The inputs into the various inference procedures were all derived by marginalizing over a single underlying distribution of belief values. For instance, to calculate B(H=T|D=T) in the prototypical example, we marginalized over the possible states where D=T (i.e. B(H=T|D=T)=B(H=T,D=T & ABC)/B(D=T &HABC)).

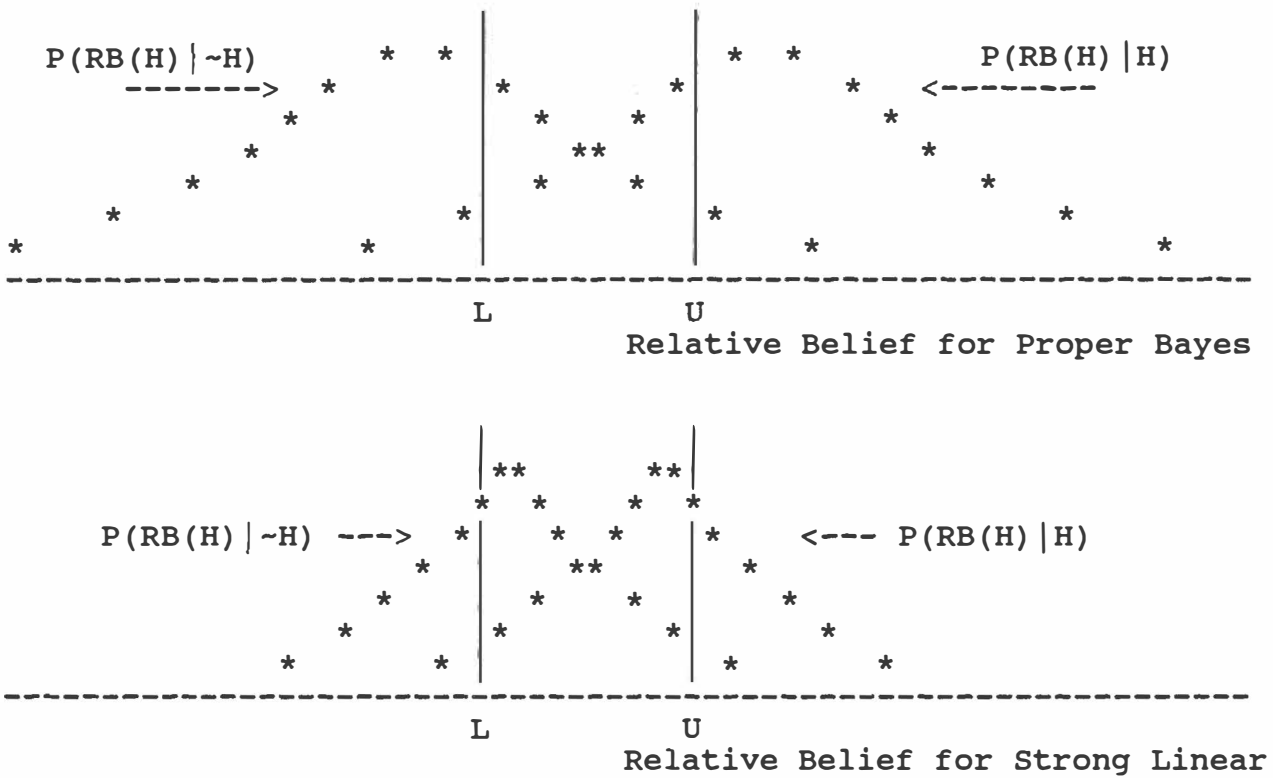

FIGURE 1
DISTRIBUTIONS OF RELATIVE BELIEF VALUES
FOR PROPER BAYES AND STRONG LINEAR

From the perspective of practical applications, this is unrealistic. It suggests that one would elicit from an agent a fully-specified probability model, accounting for all interactions, and then proceed to ignore (marginalize over) much of the elicited information to calculate the inputs needed for a simple inference procedure. The simple inference procedure does not reduce the elicitation burden. A more realistic approach is to directly assess the parameters of interest. For instance, we can let

$$B(H=T|D=T) = P(H=T|D=T) + error.$$

Using this approach we obtain a significant reduction in the elicitation burden.

The results for this variation are similar to the prototypical case. The main difference occurred for expected-RE measure where the arching effect was much less pronounced. This is shown in Table 4.

*When true probabilities do not exist* - The prototypical case assumed that for each evidential state there was a true probability that the hypothesis was true. Although convenient, this assump-tion is not required. An alternative approach to calibration error is to look at the relative frequency of truths at each level of belief. To illustrate, suppose someone solves a set of 50 hierarchical inference problems, each unique. Over the set of problems the person expresses a belief value of .6 on 100 instances. Of the set of 100 propositions, for which a belief value of .6 was expressed, what proportion were true? If the proportion was .6, then the person was well calibrated. If the proportion was .1, then the person was poorly calibrated. By setting up an error distribution for the proportion of truths at each level of belief, calibration error can be introduced without assuming that belief values are an assessment of true probability values.

TABLE 4
EXPECTED-RE RESULTS FOR DIRECT ASSESSMENT CASE

| Error Range | Simple linear | Strong linear | Naive Bayes | Strong Bayes | Proper Bayes |
|-------------|---------------|---------------|-------------|--------------|--------------|
| 0.000 | .125 | .085 | .128 | .122 | .081 |
| 0.200 | .135 | .098 | .134 | .137 | .100 |
| 0.400 | .145 | .113 | .157 | .159 | .115 |
| 0.600 | .168 | .148 | .195 | .197 | .147 |
| 0.800 | .221 | .215 | .258 | .260 | .200 |
| 1.000 | .276 | .268 | .312 | .282 | .250 |
| 1.200 | .364 | .345 | .377 | .374 | .326 |
| 1.400 | .417 | .402 | .417 | .419 | .396 |
| 1.600 | .480 | .475 | .482 | .481 | .466 |
| 1.800 | .485 | .481 | .492 | .490 | .489 |
| 2.000 | .493 | .496 | .491 | .492 | .489 |

More formally we have the following situation. For a given level of belief, say b, there is a set of pairs of expressions $\{(x_1, y_1), \ldots, (x_n, y_n)\}$ where for each pair $B(x_i|y_i) = b$. A pair is labeled true if $x_i$ & $y_i$ is true and false if $\sim x_i$ & $y_i$ is true. All pairs are either true or false.[3] Let the proportion of truths in this set be $p$. If we are willing to assume exchangability then $P(x_i|y_i \& "B(x_i|y_i) = b") = p$. Here $p - b$ is the calibration error for level b.

The results for this variation are shown in Table 5. Everything else is the same as before. Once again the results are similar to the prototypical example, although the arching effect for RE *is* a little less pronounced.

*Hierarchical Networks* - In AI, most inference systems are based on hierarchical inference models. The previous cases were all
--------------------
3. The reader may find it convenient to think of the truth value of an (xi,yi) pair in terms of counterfactuals. (xi,yi) is true on condition that if yi were true, then xi would also be true. I don't really need counterfactuals to develop this discussion, but I don't have the space here to explain how to get around them.

two levels networks with no intermediate nodes. A simple way to introduce hierarchical inference into the prototypical example is to remove one of the evidence items, but leave the structure the same. For example, we reran the prototypical case with A removed as an evidence item. The MSE and d' results are once again similar to the prototypical case, but in contrast to the previous two cases, the arching effect for RE was magnified. This is shown in Table 6.

TABLE 5
EXPECTED-RE RESULTS FOR CASE WHERE TRUE PROBABILITY
DOES NOT EXIST

| Error range | Simple linear | Strong linear | Naive Bayes | Strong Bayes | Proper Bayes |
|-------------|---------------|---------------|-------------|--------------|--------------|
| 0.000 | .122 | .089 | .117 | .119 | .081 |
| 0.500 | .159 | .126 | .174 | .176 | .138 |
| 1.000 | .284 | .260 | .308 | .309 | .262 |
| 1.500 | .425 | .421 | .436 | .440 | .404 |
| 2.000 | .487 | .485 | .495 | .497 | .464 |

TABLE 6
EXPECTED-RE RESULTS FOR HIERARCHICAL MODEL

| Error range | Simple linear | Strong linear | Naive Bayes | Strong Bayes | Proper Bayes |
|-------------|---------------|---------------|-------------|--------------|--------------|
| 0.000 | .183 | .100 | .142 | .144 | .119 |
| 0.500 | .241 | .140 | .185 | .185 | .168 |
| 1.000 | .326 | .252 | .298 | .304 | .290 |
| 1.500 | .476 | .445 | .458 | .453 | .443 |
| 2.000 | .506 | .504 | .505 | .504 | .506 |

*Other Variations* - The above results represent a small subset of the variations we have explored. These variations include alternative ways of specifying the probability of each possible state, introducing calibration error and varying thresholds. Although the numbers may change, the patterns of results shown above are generally stable.

In addition to varying the above runs, we are in the process of examining a variety of other inference procedures -- including the use of default rules, interval probabilities with Bayesian and Shaferian updates, as well as some hybrid schemes.

### 3.0 DISCUSSION

For the techniques studied here, Proper Bayes is robustly the most sensitive in discriminating true vs. false instances of a hypothesis. From the perspective of relative error, however, simply adding up the number of strong evidence items for vs. against an inference is generally safe. If this simple technique strongly supports an inference, then that inference is probably correct. Indeed, this simple inference technique is less likely to make an error in inference than a simplified Bayesian techniques and is often superior to proper Bayesian techniques.

As a final note, the reader may be curious as to how these results relate to the notion of higher order uncertainty (see Kyburg, 1988 for discussion). In particular, one might suppose that the results would have been different if we had allowed the Bayesian procedures to incorporate a probability distribution over the probability estimates. This supposition is both correct and irrelevant. To see this, consider the case where the higher-order probabilities are correct (i.e., accurately reflect the random process in the monte carlo simulation). In this circumstance, the expected probability of the hypothesis given each evidential state would then also reflect the true probability of each state -- and we are back in the zero calibration error condition. More generally the notion of probabilities of probabilities simply recurses the problem of calibration error, and doesn't really contribute to our understanding of the performance of different inference procedures.

An alternative approach to something like higher-order uncertainty is to perform a probability analysis of the characteristics of alternative uncertain inference procedures. Such an analysis can be used to identify characteristics of inference procedures that are independent of the probabilistic characteristics of an inference domain (i.e., generalizable results), as well as characteristics that seem to be context specific. This paper is an example of such an analysis.